\documentclass[conference]{IEEEtran}
\IEEEoverridecommandlockouts
\usepackage{cite}
\usepackage{amsmath,amssymb,amsfonts}
\usepackage{algorithmic}
\usepackage{graphicx}
\usepackage{textcomp}
\usepackage{xcolor}
\usepackage{multirow, makecell}
\usepackage{csquotes, bm, amsmath}
\usepackage{url}
\usepackage{fancyhdr}

\def\BibTeX{{\rm B\kern-.05em{\sc i\kern-.025em b}\kern-.08em
    T\kern-.1667em\lower.7ex\hbox{E}\kern-.125emX}}

\begin{document}

\title{Affectively Framework: \\Towards Human-like Affect-Based Agents
\thanks{This project has received funding from the Malta Council for Science and Technology through the SINO-MALTA Fund 2022, Project OPtiMaL.}
}

\author{\IEEEauthorblockN{
Matthew Barthet,
Roberto Gallotta,
Ahmed Khalifa,
Antonios Liapis,
Georgios N. Yannakakis\\
\IEEEauthorblockA{Institute of Digital Games, University of Malta, Msida, Malta.\\
Email: \{matthew.barthet, roberto.gallotta, ahmed.khalifa, antonios.liapis, georgios.yannakakis\}@um.edu.mt}}}

\maketitle
\thispagestyle{fancy}

\begin{abstract}
Game environments offer a unique opportunity for training virtual agents due to their interactive nature, which provides diverse play traces and affect labels. Despite their potential, no reinforcement learning framework incorporates human affect models as part of their observation space or reward mechanism. To address this, we present the \emph{Affectively Framework}, a set of Open-AI Gym environments that integrate affect as part of the observation space. This paper introduces the framework and its three game environments and provides baseline experiments to validate its effectiveness and potential.
\end{abstract}


\section{Introduction} 

Video games are ideal stimuli for research in Affective Computing \cite{games_hci} for several reasons. Firstly, the user is free to play in many different ways, leading to diversity in their play traces and emotional experiences \cite{affective_computing_games}. This freedom allows for deeper research into the relationship between behaviour and emotions compared to static stimuli such as videos. Games also encompass multiple modalities for modelling, such as pixels \cite{CS_Pixels}, game states \cite{again_dataset}, and controller inputs \cite{pixels_gamepad_to_affect}. Games have been at the forefront of research on reinforcement learning (RL), especially leveraging Open-AI Gym environment \cite{gym}. Indicatively, popular RL algorithms such as 
Proximal Policy Optimisation \cite{ppo} have been tested on gameplaying tasks on the Atari suite of environments \cite{gym}. While RL agents focus on playing to win \cite{yannakakis2018ai_games}, there are no environments that incorporate human affect for training emotion-aware gameplaying agents.

Training virtual agents that not only respond to game states but also human affect enables us to create a more holistic agent that can model and mimic human players better than training an agent to beat a game~\cite{go-blend}. Cutting-edge RL algorithms like Go-Explore \cite{Go-Explore} have been successfully applied to train agents that exhibit affective responses in line with target human personas \cite{go-blend} and use affect as a mechanism for beating the game \cite{affect_driven_rl}. Similar works have used RL to train affect-based virtual agents via simulated, rather than human-like, emotions \cite{simulated_emotions}, as well as intrinsic motivation to imitate human gameplay demonstrations \cite{intrinsic_rl_demonstrations}. However, we believe there is a significant hurdle for future research: developing the game environments and sourcing human gameplay and affect demonstrations is a time-consuming resource-intensive process. Without publicly available environments---and benchmarks---tied to large-scale affect corpora to reduce the development time, 
moving in this direction is bound to be a slow and uncontrollable process.

Motivated by this research gap, we present the \emph{Affectively Framework}, which incorporates a human-sourced model of arousal into its observation space. The framework follows the same philosophy as Open-AI Gym~\cite{gym} to facilitate easier integration of new games and agents later. The framework comes with three game environments based on the stimuli for the arousal video game annotation (AGAIN) dataset \cite{again_dataset}, which contains gameplay data annotated for arousal. The three game environments offer challenging gameplay as well as significantly different observation and action spaces among them. Finally, this paper presents baseline experiments on training RL agents based on both in-game behaviour and the human model of arousal, to illustrate the potential of the framework. The framework is provided as an easy-to-install package\footnote{\url{https://github.com/Matt-Barthet/Affectively-Framework}} that can be used for research into virtual agents that incorporate affect in their decision-making.

\section{Affectively Framework}

\begin{figure}[tb]
\centering
\includegraphics[width=0.9\columnwidth]{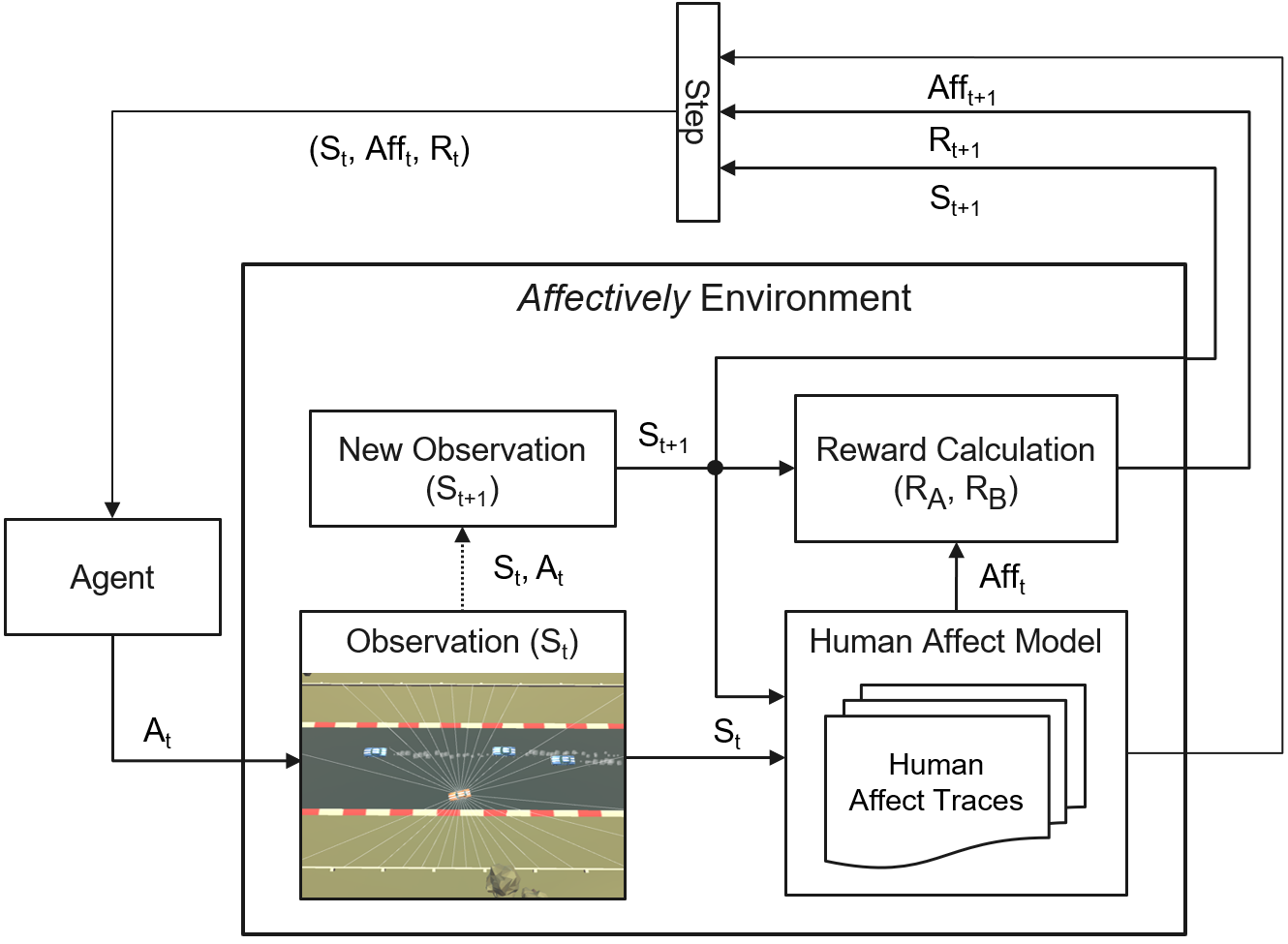}
\caption{Overview of the \textit{Affectively Framework} architecture for training affect-based RL agents. $A_t$ denotes the action taken by the agent, $S_t$ is the current observation, Aff$_t$ is the current affect value and $R_t$ is the reward assigned to the agent.} 
\label{fig:framework}
\end{figure}

The \textit{Affectively Framework} follows the RL feedback loop of Open-AI Gym~\cite{gym}, visualised in Figure \ref{fig:framework}. At each time step, the framework provides the agent with a game-state $(S_t)$ and expects an action $(A_t)$ in return. Based on the agent's action, it returns an updated game-state $(S_{t+1})$ and an \textit{environment score} $(R_E)$ that reflects the quality of the action, based on the environments' scoring system. Using $R_E$, we construct a \textit{behaviour reward} ($R_B$) for each game with additional domain knowledge in order to help the agent during training and combat the sparsity of $R_E$. The \textit{Affectively Framework} slightly modifies the typical training loop by including affect values, periodically generated every 3 seconds using a human affect model (see Section \ref{sec:affect_model}). Using the affect values generated by the human model, we provide an \textit{affect reward} ($R_A$) for training agents. In this paper, $R_A$ aims to maximise arousal, but more complex reward functions such as imitating affect \cite{affect_driven_rl, go-blend} can be used. Using these components, the total reward $R_t$ to the agent 
is calculated at each time step as per Eq.~\eqref{eq:reward_total}.  
\begin{equation}
    R_t = (1 - \lambda) \cdot n(R_B) +  \lambda \cdot R_A
\label{eq:reward_total}
\end{equation}
\noindent where $\lambda$ controls the importance of $R_B$ or $R_A$, and $n(R_B)$ is the normalised behaviour reward per game based on their extreme reward values. 
Below, we describe the game environments (Section \ref{sec:environments}) and the affect model for deriving human arousal (Section \ref{sec:affect_model}). 

\subsection{Environments} \label{sec:environments}

The \textit{Affectively Framework} includes three game environments, although it is designed for extensibility with more games in future work. The game environments span three different game genres: side-scrolling platformers (\textit{Pirates}), shooters (\textit{Heist}), and racing (\textit{Solid Rally}). These games are based on the AGAIN dataset \cite{again_dataset}, which included nine games with annotated gameplay sessions in each of them. We chose these three games for their diversity in genre, action/state space, and gameplay goals. Table \ref{tab:environment_properties} contains the important information for each game environment. 

\begin{figure}[t]
\begin{minipage}{\columnwidth}
    \centering
    \begin{tabular}{c}
        \includegraphics[width=0.48\textwidth]{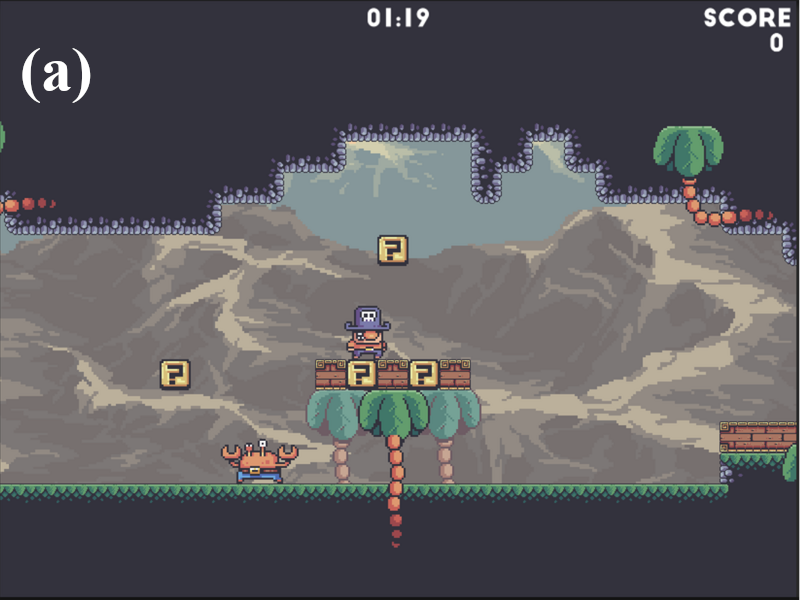}  \hspace{-5pt}
        \includegraphics[width=0.48\textwidth]{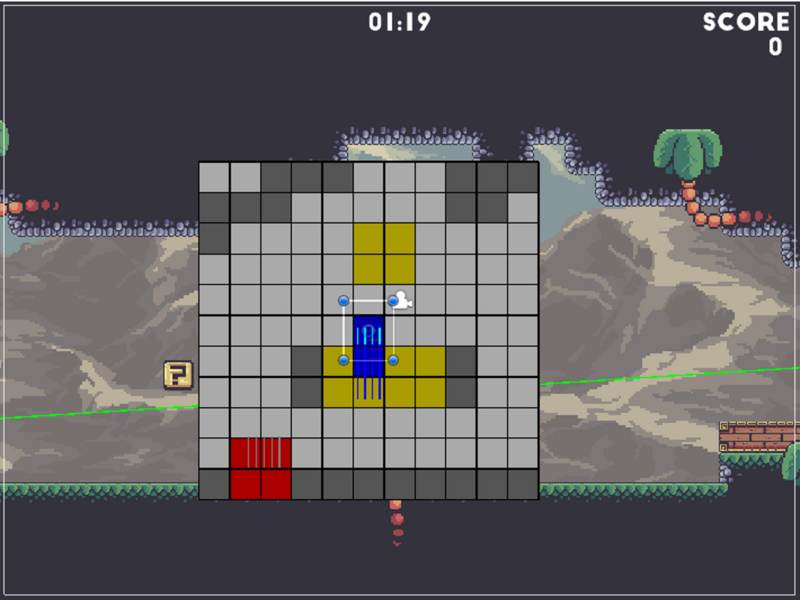}
    \end{tabular}
\end{minipage}

\begin{minipage}{\columnwidth}
    \centering
    \begin{tabular}{c}
    \includegraphics[width=0.48\textwidth]{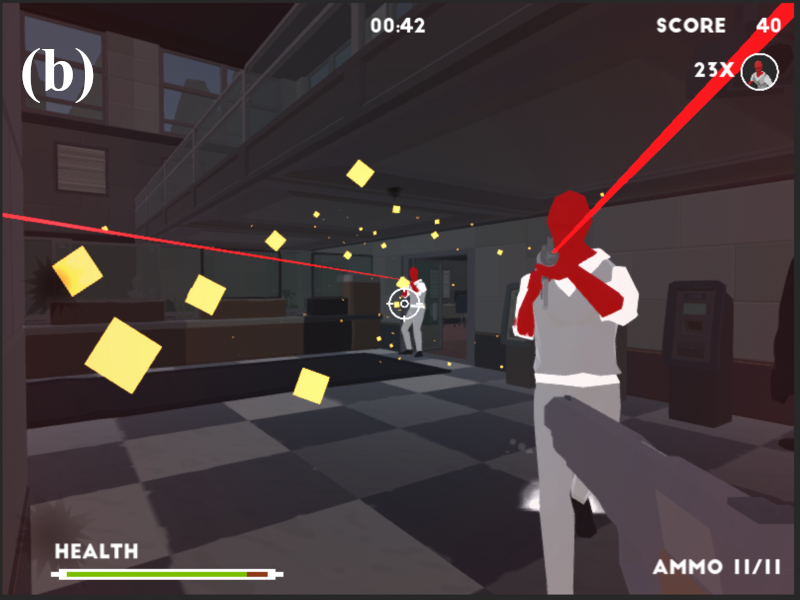} \hspace{-5pt}
    \includegraphics[width=0.48\textwidth]{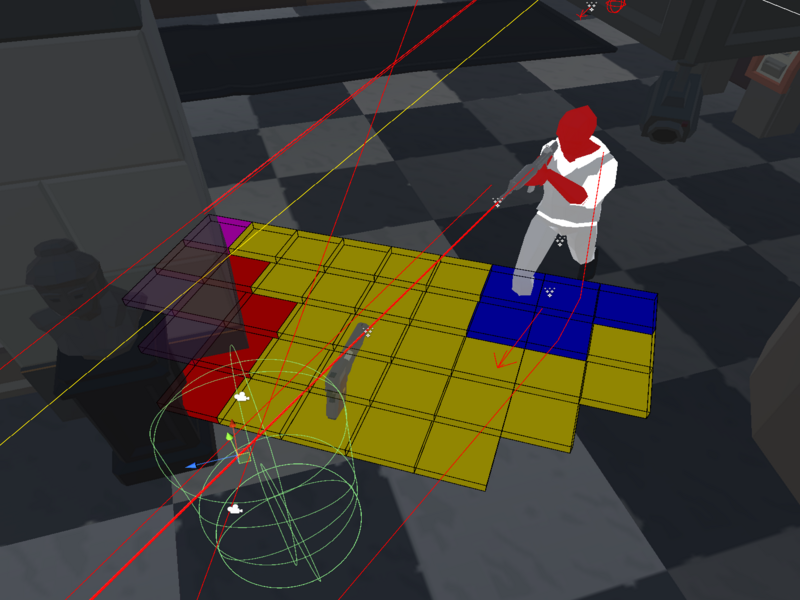}
    \end{tabular}
\end{minipage}

\begin{minipage}{\columnwidth}
    \centering
    \begin{tabular}{c}
    \includegraphics[width=0.48\textwidth]{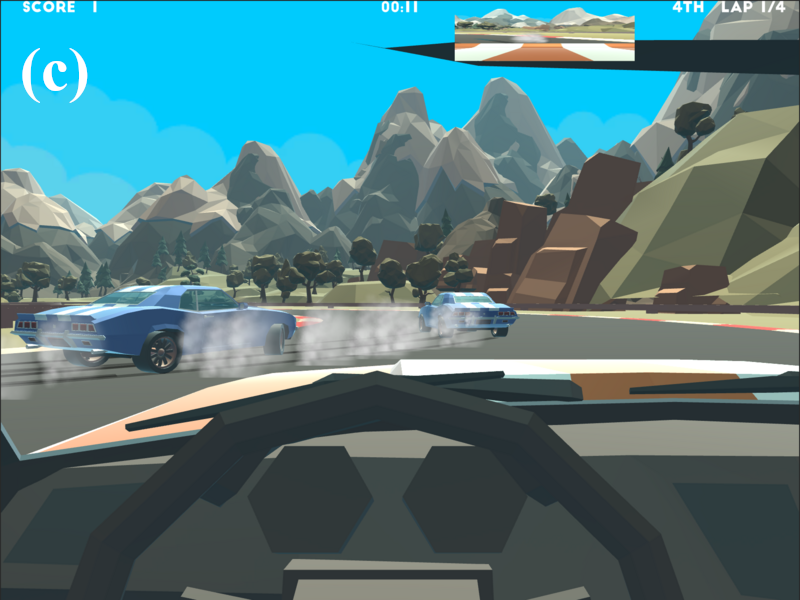} \hspace{-5pt}
    \includegraphics[width=0.48\textwidth]{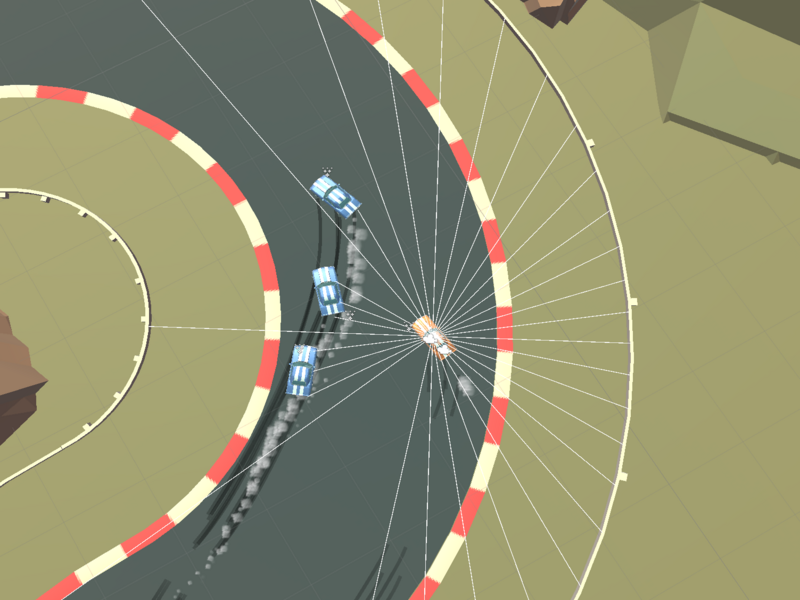}
    \end{tabular}
\end{minipage}
\begin{minipage}{\columnwidth}
    \centering
    \begin{tabular}{c}
        \begin{footnotesize}
           \hspace{-12pt} in-game screenshots \hspace{65pt} observations
        \end{footnotesize}
    \end{tabular}
\end{minipage}
\caption{The \textit{Affectively Framework}. From top to bottom: (a) \textit{Pirates} platformer game, (b) \textit{Heist} first-person shooter, and (c) \textit{Solid Rally} racing game. }
\label{fig:games}
\end{figure}

\textbf{\textit{Pirates}} is a 2D platformer heavily inspired by \textit{Super Mario Bros.} (Nintendo, 1985). In this game, the player moves from left to right and must reach a goal (exit) in under 2 minutes. During the level traversal, the player must avoid obstacles and enemies, as well as pick up coins and power-ups to improve their score. If the player dies, by falling off-screen or by colliding with an enemy, they respawn at the closest checkpoint in the level. The agent's action space consists of 2 discrete action branches. The first covers horizontal movement: the agent can move left (-1), stay still (0), or move right (1). The second covers jumping, where the agent can stay still (0) or jump (1). The environment's observation space ($S_t$) consists of an 11$\times$11 grid of integer values (see Fig.~\ref{fig:games}) corresponding to IDs of the entities within the grid, with the player always at the centre (blue). The IDs are assigned on a priority basis, with enemies (red) having the highest priority, followed by coins and power-ups (green), breakable tiles (yellow), obstacles (grey), and empty space (white). Seven additional properties are appended to $S_t$, consisting of the agent's physics variables (e.g., velocity, current direction) and game-specific properties (e.g., current health and power-up status). For \textit{Pirates}, $R_E$ is the total coins collected (10 points each) and power-ups collected (20 points each), with a maximum possible score of 460 points if the agent exhaustively searches the entire level. $R_B$ is calculated at each time step as per Eq.~\eqref{eq:reward_b_pirates}, as the change in $R_E$ from the previous frame (i.e., if the agent just picked up a coin or power-up), with two additional components: a small reward ($M_r = 0.1$) if the agent moved to the right of the screen in the last frame (to encourage level traversal), and a penalty for dying ($D=5$) to discourage reckless behaviour. 

\begin{equation}
R_B = \Delta R_E + M_r - D 
\label{eq:reward_b_pirates} 
\end{equation}

\renewcommand{\arraystretch}{1.3}
\begin{table}
    \caption{Overview of the observation and action spaces for each environment in {Affectively Framework}.}
    \centering
    \label{tab:environment_properties}
    \begin{tabular}{|c|c|c|c|}
    \hline
        \multirow{2}{*}{Environment} & $S_T$ & Discrete Actions & Continuous \\
        &   Vector Size    & (Branch Sizes) &   Actions \\
    \hline
         \textit{Pirates} & 852 & 2 (3, 2) & 0 \\
    \hline
         \textit{Heist} & 341 & 3 (3, 3, 2) & 2 \\
    \hline
        \textit{Solid Rally} & 50 & 2 (3, 3) & 0 \\
    \hline
    \end{tabular}
\end{table}

\textbf{\textit{Heist}} is a first-person shooter game where the player must explore the map and eliminate all 25 enemies in the area within the 2-minute time limit. The player has infinite ammo, but their weapon holds 11 bullets and reloads automatically 
when it is empty. If the player takes enough damage from enemies' bullets, they die and respawn at the beginning of the level; eliminated enemies do not respawn. The environment's $S_t$ consists of a $9{\times}9$ grid of IDs using the same approach described in \textit{Pirates}, this time within the field of view of the player (see Figure \ref{fig:games}). The possible IDs for tiles in the grid are obscured (i.e. out of sight, in pink), empty space (yellow), obstacles (red), and enemies (blue). In addition to the grid, the agent is supplied with a vector of 20 observations containing spatial and physics variables as well as game-specific variables such as health, ammo, and a vector to the nearest visible enemy.
The player's action space is the most complex of the three games, with 3 discrete action branches and 2 continuous actions. The first discrete action covers horizontal movement, allowing the agent to strafe left (-1), keep straight (0), or strafe right (1). The second discrete action covers depth movement, where the agent can move forward (1), stay still (0), or move backward (-1). The third discrete action covers shooting, where the player can shoot (1) or do nothing (0). The continuous actions (with values between -1 and 1) govern the horizontal and vertical movement of the camera (i.e., aiming the weapon). 
For \textit{Heist}, $R_E$ is the number of kills the player has made so far, with each kill worth 20 points. The maximum possible score is 500 points ($20 \times 25$ enemies). At each time step $R_B$ is calculated as per Eq.~\eqref{eq:reward_b_heist}, as the changes in $R_E$ from the previous time step (i.e., if the agent has just killed an enemy), with two additional components: a small exploratory reward ($E=1$) every time the agent enters a new area of the map (the map is partitioned a priori into a grid of $5{\times}5{\times}5$ cubes) and the inverted angle ($A$) between the agent and the nearest enemy, normalised between -1 (facing away from enemy) and 1 (facing the enemy). In other words, the agent is rewarded for exploring new areas of the map and for facing nearby enemies.

\begin{equation}
R_B = \Delta R_E + E + A
\label{eq:reward_b_heist} 
\end{equation}

\textbf{\textit{Solid Rally}} is a first-person 3D racing game where the player controls a rally car and must finish a 3-lap race ahead of the three opponents' AI-controlled cars within 2 minutes. The track contains 8 waypoints that the player must drive through, which act as reset points if their car gets stuck. The environment's $S_t$ consists of a vector of 50 real values, which include various physics variables (e.g. velocity, rotation) as well as spatial variables (e.g. distances and angle to next waypoint, distance to the nearest objects surrounding the car). The action space consists of two discrete action branches: one for steering the car left (-1), right (1) or straight (0), and one for forward movement, accelerating forward (1), braking/reversing (-1), or coasting (0). For \textit{Solid Rally}, $R_E$ is the number of waypoints the agent has driven through so far, with a maximum possible score of 24 ($8 \times 3$ laps). At each time step, $R_B$ is calculated as per Eq.~\eqref{eq:reward_b_solid} as the change in $R_E$ compared to the previous state (i.e. if the agent has just driven through a waypoint), with an additional component: the product of the agent's speed ($S$) and inverted angle ($A$) to the nearest checkpoint, normalised between 0 (facing away) and 1 (facing directly toward). These additional components encourage agents to drive fast and in the right direction. 

\begin{equation}
R_B=\Delta R_E + (S \cdot A)
\label{eq:reward_b_solid}
\end{equation}

\renewcommand{\arraystretch}{1.3}
\begin{table*}[t]
\centering
\caption{Results for RL agents and baselines tested on three different reward functions, averaged across 30 evaluation runs and include the 95\% confidence interval. Bold results correspond to the best reward observed for each column. }
\label{tab:baselines}
\begin{tabular}{r|c|c|c|c|c|c|c|}
\cline{2-8}
& \multirow{2}{*}{\makecell{\textbf{Agent}}} & \multicolumn{2}{c|}{\textbf{Pirates}} & \multicolumn{2}{c|}{\textbf{Heist}} & \multicolumn{2}{c|}{\textbf{Solid Rally}} \\
\cline{3-8}
&  & Final $R_E$ & $\Bar{R_A}$ & Final $R_E$ & $\Bar{R_A}$ & Final $R_E$ & $\Bar{R_A}$ \\
\cline{2-8}
\noalign{\vskip 1mm} 
\cline{2-8}
& \makecell{Random} & $0.080\pm0.013$ & $0.578\pm0.019$ & $0.027\pm0.013$ & $0.594\pm0.027$ & $0.007\pm0.006$ & $0.397\pm0.013$ \\
\cline{2-8}
& \makecell{Humans} & $\bm{0.476\pm0.036}$ & $0.501\pm0.038$ & $\bm{0.461\pm0.039}$ & $0.496\pm0.036$ & $\bm{0.795\pm0.029}$ & $0.496\pm0.038$ \\
\cline{2-8}
\noalign{\vskip 1mm}
\cline{2-8}
\parbox[t]{2mm}{\multirow{3}{*}{\rotatebox[origin=c]{90}{\textbf{PPO}}}} & \makecell{Max. Behaviour} & $0.278\pm0.031$ & $0.508\pm0.016$& $0.108\pm0.009$ & $0.518\pm0.021$ & $0.413\pm0.052$ & $0.4157\pm0.034$ \\
\cline{2-8}
& \makecell{Blended} & $0.000\pm0.000$ & $0.827\pm0.010$ & $0.045\pm0.002$ & $0.581\pm0.021$ & $0.653\pm0.052$ & $0.385\pm0.028$ \\
\cline{2-8}
& \makecell{Max. Arousal} & $0.000\pm0.000$ & $\bm{0.863\pm0.012}$ & $0.000\pm0.000$ & $\bm{0.850\pm0.010}$ & $0.000\pm0.000$ & $\bm{0.592\pm0.017}$ \\
\cline{2-8}
\end{tabular}
\end{table*}

\subsection{Affect Model} \label{sec:affect_model}
The game environments selected for the \emph{Affectively Framework} were originally built for the AGAIN dataset, where each environment comes with over 122 human game sessions annotated in a first-person manner in terms of arousal \cite{again_dataset}. Our framework uses the same affect model for all the environments, which is based on arousal transitions within the corpus of human affect annotations in similar game-state transitions as the ones encountered by the agent at this point. The affect model is built using the k-nearest neighbours algorithm \cite{knn} (KNN), with $k=5$. We follow the same methodology for deriving this affect model across all three games, based on successful arousal modelling attempts when the dataset was collected \cite{again_dataset}. For each environment, every annotated play trace is processed as consecutive 3-second time windows: each entry contains the mean arousal value of that time window along with a feature vector $\vec{P}$ of game-specific variables such as car speed, player health, and current score. These variables $\vec{P}$ are used in AGAIN \cite{again_dataset} and are different from the game-state variables ($S_t$) described in Section \ref{sec:environments}.
We construct pairs of consecutive time windows from these gameplay sessions, and label affect in terms of decreasing, increasing, or stable between the two time windows. As with previous studies \cite{again_dataset}, we discard stable arousal data points from the corpus.

During execution, the affect model calculates the agent's current (and previous) feature vector $\vec{P}$ every 3 seconds of in-game time (via averaging). It then queries the dataset for the 5 nearest neighbours using an inverse-distance weighted averaging (i.e., the closer the neighbour is to the current state, the more influence it has on the output) as used in previous studies \cite{go-blend}. This means that the model finds the closest transitions (in terms of 3-second time windows) made by humans in the corpus and produces a weighted average of their arousal transition (increase or decrease) weighted by the $\vec{P}$ distance. The result is a value between 0 (all agree on decrease) and 1 (all agree on increase). We test the same affect reward ($R_A$) in all three environments, which is to maximise arousal (rewarding state transitions that increase arousal) generated by the KNN model at each time window. It is important to note that since the model generates values every 3 seconds, $Aff_t=0$ for any time steps in between.

\section{Experimental Protocol} \label{sec:protocol}
In this paper, we evaluate the \textit{Affectively Framework} by training RL agents to maximise in-game performance, arousal, or a combination thereof. For each evaluation, we report the final $R_E$ and the $\Bar{R_A}$ of their arousal trace taken at the end of the session. We detail the baselines (random agent and human demonstrations) and the RL agent that uses Proximal Policy Optimisation (PPO) below:

\subsubsection{Random Agent}
This baseline agent takes random actions by uniformly sampling the environment's action space. Results are averaged from 30 runs on the same environment.

\subsubsection{Humans}
We use humans from the AGAIN corpus as a baseline to assess both their performance (in terms of in-game score) and their arousal levels (in terms of increases or decreases of their own annotation trace in 3-second consecutive time windows) throughout the gameplay session. The human sessions are identical to the agents in terms of both scoring and ending conditions (2 minute time limit).

\subsubsection{PPO Agent}
The PPO algorithm \cite{ppo} is a very popular RL algorithm implemented in the Stable Baselines (SB) library \cite{stable_baselines}. This agent uses the default parameters provided by SB. For each scenario, we train a single agent for 1 million time steps, before evaluating them 30 times with exploration turned off (i.e., always picking the best action). The agent is evaluated using SB's stochastic action prediction to sample actions from the network. The PPO agents use three different rewards:
\begin{itemize}
    \item \textbf{Max. Behaviour} which uses as reward $R_t$ of Eq.~\eqref{eq:reward_total} with $\lambda=0$, thus optimising only for $R_B$. We expect this agent to play only to win (i.e., maximise their $R_E$).
    \item \textbf{Max. Arousal} which uses as reward $R_t$ of  Eq.~\eqref{eq:reward_total} with $\lambda=1$, thus optimising only for $R_A$. We expect this agent to attempt to maximise the arousal score derived from human examples, which may lead to unexpected and uncontrollable behaviours.
    \item \textbf{Blended} which uses as reward $R_t$ of  Eq.~\eqref{eq:reward_total} with $\lambda=0.5$, optimising for $R_B$ and $R_A$ equally. This agent is the most interesting, as it is expected to act in a more human-like fashion, assuming a human plays to win and to enjoy themselves (via an increased arousal state).
\end{itemize}

\section{Results}
Table \ref{tab:baselines} shows the final in-game score ($R_E$) normalised to $[0,1]$ based on the maximum possible score per game (see Section \ref{sec:environments}), as well as the $\Bar{R_A}$ for the PPO agents and the baselines described in Section \ref{sec:protocol}. As expected, a random agent is not capable of effectively playing any of the three environments, achieving very low scores ($R_E$) in all 30 evaluation runs. When it comes to affect, random agents did not exhibit any consistent pattern across games: agents would get stuck very quickly and generate the same arousal values repeatedly until the end of the episode.  

The PPO agents, even when rewarded only based on $R_B$, played every game worse than the humans in the AGAIN corpus and reached much lower final scores ($R_E$). This can be somewhat expected from the short training times. \textit{Heist} proved to be particularly challenging for the agent, likely due to its complex action space (with 2 continuous actions) and challenging gameplay (requiring both way-finding and aiming). While the final scores of the $R_B$-based agent were approximately half those of the average human for \textit{Solid Rally} and \textit{Pirates}, the agents' behaviour was robust. However, for \textit{Pirates} the agent tended to rush to complete the level quickly and sometimes ignore collectables, likely due to the (constant) $M_r$ component of $R_B$ in Eq.~\eqref{eq:reward_b_pirates} that rewards moving right. For \textit{Solid Rally}, where moving forward quickly was the only type of desired gameplay, this sufficed to perform well.

When PPO agents were trained to maximise arousal ($R_A$), the agents significantly improved their average arousal values, surpassing the random agent and the human demonstrations alike. However, the behaviour of these agents was very poor across games (with $R_E=0$ in all 30 runs and all three games). We observed that the agents would quickly find a region of the environment close to the start with high arousal and exploit that area, rather than explore the environment. This illustrates the deceptive reward space when rewarding affect alone. Using an RL agent which supports and rewards exploration, such as Go-Explore \cite{Go-Explore}, will likely improve this PPO agent's behaviour---as seen in previous studies \cite{go-blend}.

PPO agents trained on blended reward showed an inconsistent trend across game environments. In \textit{Solid Rally}, the blended agent managed to improve upon its final score compared to the pure behaviour optimiser, but had the worst arousal score. In this case, the behaviour reward overpowered the affect reward given its richer and less deceptive nature. The opposite was true in \textit{Pirates}, which featured large increases in arousal at the very start of the game. The affect reward ($R_A$) thus overpowered the behaviour reward ($R_B$), leading the agent to stay largely still---similar to the arousal optimising PPO agent. Parsing the training data, we observed that the agent initially explored deeper into the level, achieving the best $R_E$ score of $0.106$ in the first 50 episodes, but quickly converged on exploiting the starting area of the level which exhibits high arousal in the annotation traces. While we observed a more balanced outcome (in terms of $R_E$ and $R_A$) in \textit{Heist}, the agent performed poorly compared humans' $R_E$ and only marginally better than the random agent. We believe that the game proved too difficult for the agent to learn an effective policy within the limited training time provided.

\section{Discussion}
Experiments with the built-in agents and reward systems provided by the \emph{Affectively Framework} highlighted the inherent challenge of optimising affect in gameplaying agents. Affect, as derived from similar game state transitions in a corpus of annotated human gameplay traces, can be both deceptive and overpowering as an RL reward. In our results, we found that balancing $R_A$ and $R_B$ in the blended reward function is a significant challenge for RL algorithms such as PPO, and requires more tuning. We acknowledge that a major limitation is the sparsity of $R_A$ rewards, which are provided once every 3 seconds (and are not triggered by a specific in-game event apart from the game clock); this may confuse the agent, especially for blended reward schemes. Future work should explore alternatives, such as a moving time window for deriving agents' parameter vector to match human demonstrations (see Section \ref{sec:affect_model}), or copying the last calculated $R_A$ reward into the reward scheme of Eq.~\eqref{eq:reward_total}. 

Moreover, the size of AGAIN's corpus allows us to find more relevant neighbours for unseen game-state transitions made by the agent. However, using only a subset of these gameplay traces could lead to better behaviours. In our previous study on \textit{Solid Rally} \cite{go-blend}, clustering play traces into four clusters allowed us to train more high-performing agents based on expert players alone, using a blended reward. We plan to enhance the \textit{Affectively Framework} with pre-made clusters of players as part of all three environments, so that future agents can be trained with a particular target persona in mind (e.g. imitate ``Expert'' behaviour and arousal). 

A straightforward area for improvement is to expand the current experiments with longer training times and hyper-parameter tuning, in order to ensure optimal performance of the PPO algorithm. More experiments with different RL algorithms and affect rewards, especially in light of the poor behaviour of PPO agents that maximise arousal, are necessary. Leveraging the \textit{Affectively Framework} for research into RL agents in multiple games, including our previous work on Go-Blend \cite{go-blend} in \textit{Solid Rally}, is a major motivation for this software release. Value-based RL algorithms, such as deep RL \cite{deep_rl_atari}, are also interesting additions to our baseline agents. Due to the complexity of the action spaces (especially in \textit{Heist}), integrating deep RL would require some work. Finally, an interesting direction for future research is to use a single observation space (e.g. using the screen's pixels) across games to enable more in-depth studies on the generalisability of RL agents. 

\section{Conclusion}
This paper introduced the \textit{Affectively Framework}, a framework compatible with the standard Open-AI Gym API~\cite{gym} for training game-playing agents that not only care about behaviour but also affect. We introduced three game environments and tested them with two baseline agents (a random agent and a PPO agent) in various behaviour and affect tasks. By laying the groundwork for future research through an easy-to-use software package and test bed environments, future studies can shed light on the relationship between player behaviour and their affect response during gameplay.

\section*{Ethical Impact Statement}
This paper makes use of an existing dataset \cite{again_dataset} of human demonstrations collected from crowd workers on the Amazon Mechanical Turk platform. This dataset is publicly available, and participants gave their consent for their data to be stored and utilised in an anonymous fashion. To the best of our knowledge, there is no significantly negative application of the methods we use in this paper and no added privacy or discrimination risk. The data used in the paper and environments introduced are publicly available for scientific reproducibility and for extensions of this study.

\bibliographystyle{ieeetr}
\bibliography{bibliography}

\end{document}